\documentclass{article}

\usepackage{arxiv}

\usepackage[utf8]{inputenc} 
\usepackage[T1]{fontenc}    
\usepackage{hyperref}       
\usepackage{url}            
\usepackage{booktabs}       
\usepackage{amsfonts}       
\usepackage{nicefrac}       
\usepackage{microtype}      
\usepackage{graphicx}
\usepackage{natbib}
\usepackage{doi}
\usepackage{bbm}
\usepackage{bm}

\usepackage{amsmath}
\usepackage{amssymb}
\usepackage{amsfonts}
\usepackage{listings}
\usepackage[dvipsnames,svgnames,x11names]{xcolor}
\usepackage{placeins}

\usepackage{todonotes}
\usepackage{fancyhdr}
\usepackage{cleveref}

\usepackage{pgfplots}   
\usepackage{tikz}

\graphicspath{{./Figures/}}
\pgfplotsset{compat=1.11}
\newlength\fwidth

\usepackage{dsfont}

\usepackage{subcaption}

\newcommand{\R}{\mathbb{R}}

\usepackage{authblk}
\usepackage[normalem]{ulem}

\definecolor{codegreen}{rgb}{0,0.6,0}
\definecolor{codegray}{rgb}{0.5,0.5,0.5}
\definecolor{codepurple}{rgb}{0.58,0,0.82}
\definecolor{backcolour}{rgb}{0.95,0.95,0.92}

\lstdefinestyle{mystyle}{
    backgroundcolor=\color{backcolour},   
    commentstyle=\color{codegreen},
    keywordstyle=\color{blue},
    numberstyle=\tiny\color{codegray},
    stringstyle=\color{codepurple},
    basicstyle=\ttfamily\footnotesize,
    breakatwhitespace=false,         
    breaklines=true,                 
    captionpos=b,                    
    keepspaces=true,                 
    numbers=left,                    
    numbersep=5pt,                  
    showspaces=false,                
    showstringspaces=false,
    showtabs=false,                  
    tabsize=2
}

\title{Distributional Regression with Tabular Foundation Models: Evaluating Probabilistic Predictions via Proper Scoring Rules}
\author[1]{Jonas Landsgesell \thanks{jonaslandsgesell\_at\_gmail dot com}}
\author[1]{Pascal Knoll \thanks{knollpascal00\_at\_gmail dot com}}
\author[2,3]{Tizian Wenzel \thanks{wenzel\_at\_math dot lmu dot de}}
\affil[1]{University of Stuttgart (Stuttgart, Germany)}
\affil[2]{Ludwig Maximilian University of Munich (Munich, Germany)}
\affil[3]{Munich Center for Machine Learning (Munich, Germany)}
\date{February 27, 2026}	

\hypersetup{
pdfauthor={Jonas Landsgesell, Pascal Knoll, Tizian Wenzel},
pdfkeywords={Fine-tuning, Distributional Regression, Benchmarking, Proper Scoring Rules},
}

\begin{document}
\maketitle

\begin{abstract}
We study how the objective used to train or fine-tune a tabular foundation model affects its distributional predictions.
Models such as TabPFN and TabICL provide forecasts of the full conditional distribution, but regression benchmarks including TabArena and TALENT assess them mainly through RMSE and $R^2$.
Such assessments reward accurate conditional means while leaving the remaining distributional information unassessed.
We therefore examine proper scoring rules as objectives for training and fine-tuning and as criteria for evaluating and ranking these models.
Every strictly proper rule favors the true distribution at the population level,
however, this shared optimum does not determine how learning proceeds with a finite sample.
Our examples illustrate score-dependent preferences between forecasts, and our fine-tuning experiments show differences in predictive performance under different objectives.
Experiments that fine-tune TabPFN and TabICL on several datasets reveal that changing the training objective can change performance across evaluation scores, 
with non-trivial interactions between the two.
By connecting established methods from distributional regression, scoring-rule theory, and foundation-model adaptation, we show that selecting the score is part of specifying the model and can measurably alter the distribution it learns.
\end{abstract}


\section{Introduction}

Tabular foundation models (TFMs) and prior-data fitted networks (PFNs) \citep{hollmann2022tabpfn, hollmann2025accurate, qu2025tabicl, qu2026tabiclv2} have significantly advanced predictive performance on small-to-medium scale benchmarks, surpassing previously dominant gradient-boosted models.
These TFMs leverage the concept of in-context learning (ICL) via the attention mechanism \citep{vaswani2017attention} during the forward pass of the neural network: the model is provided with in-context examples $\{(\bm{x}_i, y_i)\}_{i=1}^n$ and uses them to make predictions without updating the model weights. Therefore, ICL is a paradigm shift from the traditional train-then-inference approach, very similar to the emergence of promptable large language models.
ICL improves sample efficiency and generalization during extrapolation \citep{zhang2026mitra}.

While TabPFN and TabICL introduced discretized distributional regression early on, other TFMs like Limix \citep{zhang2025limix} and Mitra \citep{zhang2026mitra} initially predicted only point estimates before adopting distributional outputs in later iterations (after time of writing).

\begin{figure}[htbp]
\centering
\includegraphics[width=0.9\textwidth]{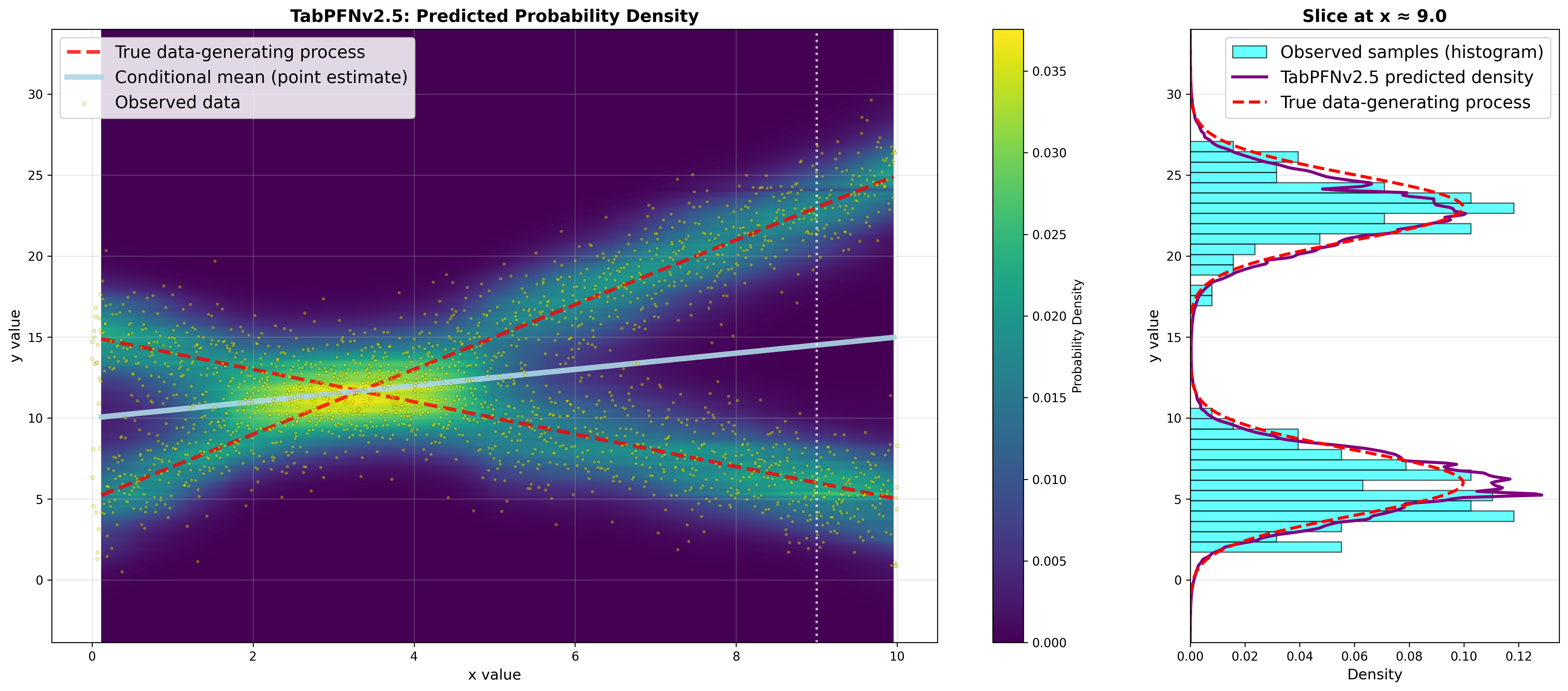}
\caption{Toy example that illustrates the downside of point based estimators.
Left: Bimodal data set, where the point estimate of the mean value (light blue) is typically far from the observed data points, that are scattered around a data generating process (DGP). The color coding in the heatmap represents the predicted probability density provided by TabPFNv2.5 which describes the distribution of the samples well.
Right, detailed view for a slice at $x \approx 9.0$ displaying a histogram of the samples (cyan), the ground truth data generating process (red) and the probability density prediction of TabPFN (purple).}
\label{fig:x-shape}
\end{figure}

Performing distributional regression produces full probabilistic forecasts. 
Practically, this means that the models either output predictive densities or predictive cumulative distribution functions.
In practice, the \textit{representation} of the forecast may vary from one model to another, e.g., a forecast can be represented as a discretized density (i.e., a probability mass function or histogram, \Cref{fig:x-shape}), as in TabPFN, or as a set of quantiles, as in TabICLv2.
Performing distributional regression is a major shift away from point estimates which were previously elicited with certain scoring functions \citep{De_Finetti1975, gneiting2011making} (MSE for the mean, MAE for the median, pinball loss for quantiles).

Point estimates are valid, well-understood targets that come with their own guarantees; the issue we address is therefore not the point estimate itself, but the naive \texttt{fit}/\texttt{predict} workflow common in machine learning, in which a single functional is reported without checking whether it adequately describes the conditional distribution.
Statisticians have long been aware of this and address it through an iterative modeling cycle (fit, diagnose via residual plots, respecify), yet in practice often only a mean value is estimated even when the problem calls for richer modeling \citep{KNEIB202399} -- a tendency already noted by Galton~(1889): ``statisticians commonly limit their inquiries to averages, and do not revel in more comprehensive views'' \citep{KNEIB202399}.
In modern machine learning pipelines, however, the black-box nature of the models together with the scale and heterogeneity of the data types they are applied to can make this manual diagnose-and-respecify loop cumbersome. This motivates models that are \emph{forgiving} in the sense of capturing the full conditional distribution out of the box, without a per-dataset modeling cycle involving many specialized methods.

We illustrate the failure of the habitual practice of applying a \texttt{fit}/\texttt{predict} mean estimator with a toy dataset in Figure \ref{fig:x-shape} where the target variable has multiple modes\footnote{One might argue that such a bimodal response may be due to an unobserved confounder. However, one has to deal with the data at hand, and therefore modeling the aleatoric (irreducible) uncertainty is important to make better decisions.}.
The heat map (left plot of \Cref{fig:x-shape}) compares the distributional forecast
of TabPFN v2.5
to the conditional mean estimate.
While the distributional prediction of the tabular foundation model closely approximates the target distribution in this example,
even a perfect mean value prediction does not reflect the underlying variability.
In fact, the mean value prediction may lie far from any observed realizations, 
potentially rendering it useless depending on the utility.
Furthermore, a standard symmetric prediction interval around the mean would cover empty space without samples, whereas a distribution-aware summary such as the highest density interval (HDI), or the conditional histogram shown in the right plot of \Cref{fig:x-shape}, more closely captures the underlying distribution.

\setlength\fwidth{.45\textwidth}

Given a probabilistic forecast, it is, however, natural to ask the following fundamental question \citep{dawid1984present, gneiting2007calibrationandsharpness}: 
\begin{center}
What constitutes a ``good'' probabilistic forecast during training and inference? \\
\end{center}
The prequential principle makes realized outcomes the basis for assessing predictions \citep{gneiting2007calibrationandsharpness, dawid1984present, philip1999prequential}.
Proper scoring rules implement this principle for probabilistic forecasts \citep{gneiting2007calibrationandsharpness, gneiting2007strictly, gneiting2014probabilistic}.
Propriety ensures that the true distribution minimizes the expected score, but this population statement does not make the available training objectives interchangeable.
With finitely many observations, the selected objective shapes optimization and the errors that a fitted model prioritizes \citep{merkle2013choosing, waghmare2025proper, wessel2025enforcing}.
Assessment under another score can consequently give a different ordering of models \citep{merkle2013choosing, buchweitz2025asymmetric, landsgesell2026scoringbench}; we discuss this distinction in \Cref{sec:choice} and examine its consequences through the fine-tuning experiments in \Cref{sec:results}.

Building on established proper scoring rule theory \citep{gneiting2007strictly, gneiting2011comparing, gneiting2011making}, this paper shows how to apply its concepts in practice to modern tabular foundation models.
In particular, we use proper scoring rules such as for example the continuous ranked probability score (CRPS) as fine-tuning objectives and as evaluation metrics for predictive distributions.
The particular choice of scoring rule should be aligned with the decision-making context and underlying risk structure \citep{carvalho2016overview} in the evaluation.
However, during model development, the choice of the scoring rule for training and fine-tuning can be treated as a hyperparameter. With limited data and pre-trained models, we observe that different scoring rules yield different performance outcomes and are not necessarily self-aligned (see \Cref{sec:results}).

Since different scoring rules penalize errors in probabilistic forecasts differently (a key distinction under finite data or limited model capacity), this choice has measurable consequences for model rankings. Practitioners can often narrow down appropriate rules using domain knowledge (e.g., whether tail or center performance matters, or whether a local scoring rule is needed \cite{du2021beyond}), though guiding this choice remains a hard problem and is beyond this paper's scope.

\subsection{Related Work}

\subsubsection{Probabilistic Forecasting or Distributional Regression}

Distributional regression advances naive modelling by modeling the entire conditional distribution $F(Y=y \mid \mathbf{x}) = P(Y \leq y \mid \mathbf{X} = \mathbf{x})$ rather than a single functional of it, thereby providing a complete characterization of uncertainty \citep{KNEIB202399, henzi2021isotonic}. 
Many approaches to distributional regression exist \citep{KNEIB202399, gneiting2007calibrationandsharpness}, 
including quantile regression, Bayesian methods producing posterior predictive distributions, and Monte Carlo methods (sample based). 
Recently, conformal predictive systems have also been proposed to provide distributional predictions with specific finite sample guarantees \citep{vovk2017nonparametric, vovk2018cross, bostrom2021mondrian}.

The goal of a distributional regression method is to approximate the (conditional) distribution $F$ of the ground truth generating process with a predictive distribution $\hat{F}(Y=y \mid \mathbf{x}) \in \mathcal{F}$, where $\mathcal{F}$ denotes a suitable family of distributions over $\mathbb{R}$. 
This shifts the modeling question from \emph{``what is the mean value of $Y$?''} to \emph{``How probable are certain outcomes of $Y$ and how good is this predicted distribution''} \citep{dawid1984present, gneiting2007calibrationandsharpness}. 
Distributional regression, therefore, strictly generalizes classical regression, and the only cost is a more demanding modeling and evaluation framework but potentially much richer predictions that can be used for better decision making \citep{lindley2013understanding, gneiting2007calibrationandsharpness, vovk2018cross}.
While conditional density estimation \citep{izbicki2016nonparametric} targets $p(Y|X)$, optimum score estimation \citep{waghmare2025proper, gneiting2014probabilistic, dawid2016minimum} can target either the conditional or even the joint density $p(Y,X)$.
Deciding what constitutes a good probabilistic forecast is a fundamental question~\cite{dawid1984present} that we address with proper scoring rules in the following sections.

\subsubsection{Proper scoring rules for probabilistic forecasting.}
\label{subsec:related_work_proper_scoring_rules}

Proper scoring rules are a fundamental tool for evaluating and ranking according to the quality of probabilistic forecasts in terms of calibration and sharpness \citep{gneiting2007strictly}.
There is a long tradition of evaluating probabilistic predictions via proper scoring rules. 
Much theoretical work about the evaluation of probabilistic predictions via proper scoring rules was described by \citet{gneiting2007strictly}.
\cite{dawid2007geometry} discusses how scoring rules and decision problems interact and how a ``decision geometry'' is formed.
\citet{gneiting2011comparing} introduced threshold- and quantile-weighted scoring rules to accommodate asymmetric risk structures, providing the theoretical basis for the weighted CRPS. 
\citet{waghmare2025proper} explain how learning from finite samples depends on the selected score, distinguishing a shared population optimum from the behavior of gradient-based optimization.
The implications extend to model comparison: in probabilistic classification, \citet{merkle2013choosing} found that changing the evaluation score can change the ordering of competing models.
The direction of a forecast error can also affect its penalty.
\citet{buchweitz2025asymmetric} establish this asymmetry between under- and overestimation for a broad family that contains the CRPS and the logarithmic, quadratic, and energy scores.
\citet{henzi2021isotonic} and \citet{gneiting2007strictly} therefore argue that ``distributional regression techniques should be compared and evaluated using proper scoring rules.''\\
The literature on scoring rules establishes that the choice of evaluation and training criterion is theoretically non-trivial and has measurable consequences for model rankings and inductive bias.
We investigate how choosing among proper scoring rules affects the fine-tuning and evaluation of tabular foundation models.

\subsubsection{Tabular foundation models.} 

There used to be an active debate in the literature about
whether neural networks are able to outperform the gold standard of gradient-boosted decision trees on tabular data \citep{mcelfresh2023neural}.
The recent emergence of tabular foundation models (TFMs), including PFN-based approaches such as TabPFN \citep{hollmann2022tabpfn, hollmann2025accurate} and large-scale pretrained models such as TabICLv2 \citep{qu2026tabiclv2}, 
represents a fundamental shift in tabular data learning. 
TFMs now frequently outperform classical gradient-boosted decision trees (GBDTs) on small-to-medium-scale benchmarks. 
Unlike the conventional train-then-inference paradigm, 
TFMs reframe supervised learning as an in-context learning (ICL) task.
Instead of updating parameters during training, the model infers task structure and inductive biases directly from the dataset at inference.
\citet{koshil2025context} illustrates that in-context learning can be understood and achieved with soft-$k$-nearest neighbor architectures with a learned distance function. \\
We finally remark that TabPFN and TabICL enable full-distributional outputs by design,
representing a qualitative departure from point-estimate methods.
The benchmarks discussed here assess regression primarily through point-prediction metrics \citep{landsgesell2026scoringbench}.
We therefore propose proper scoring rules as the natural evaluation framework for these models, see \Cref{sec:choice}.
Beyond evaluation, the rich forecasting literature on combining probabilistic forecasts \citep{gneiting2013combining, tibshiraniforecast2023} 
offers directly relevant methodology for TFMs that rely on ensembling \citep{hollmann2025accurate, qu2026tabiclv2}.

\subsubsection{Tabular Benchmarks.}

TALENT \citep{liu2025talent}, TabArena \citep{erickson2025tabarena}, and TabReD \cite{rubachev2024tabred} provide widely used comparisons of tabular learning methods.
Their regression assessments generally rely on errors of point predictions, including RMSE and $R^2$, and therefore leave the models' distributional forecasts unassessed.
Recent work has explored fine-tuning these models:
\citet{tanna2025tabtune} introduced TabTune, a unified library for inference and fine-tuning of tabular foundation models, and \citet{tanna2026exploring} systematically compared meta-learning, supervised fine-tuning, and parameter-efficient fine-tuning (PEFT) strategies. 
These studies establish that fine-tuning can be beneficial in the evaluated settings, but they focus exclusively on \emph{whether} and \emph{how} to fine-tune in terms of parameter update strategies, treating the training loss as fixed. 
Our comparison additionally varies the scoring rule used for fine-tuning.
\cite{izbicki2026benchmarking} in a concurrent work described a benchmark for continuous density estimation (CDE),
where the CDE loss is used as a metric for evaluation. 
The CDE loss is derived from the continuous Brier score \citep{rudemo1982empirical} and sometimes called quadratic score \citep{gneiting2007strictly}.

We concurrently proposed a benchmark for distributional regression \cite{landsgesell2026scoringbench}, 
where we evaluate the performance of different scoring rules for training and evaluation across the ScoringBench regression datasets (which also stem from OpenML \citep{bischl2025openml}).

\subsection{Contributions}
\label{subsec:contributions}

The main contributions are summarized below:
\begin{itemize}
\item \textbf{Evaluation mismatch for distributional regression in tabular benchmarks.}  
This paper highlights that point-estimate metrics (RMSE, $R^2$) are insufficient when models already produce full predictive distributions.  
Using a toy model (\Cref{fig:x-shape}), we motivate the need to evaluate and optimize probabilistic forecasts via proper scoring rules  
rather than rely solely on conditional means or other point functionals.
\item \textbf{Scoring-rule choice in model training.}
We illustrate how scoring rules can change optimization updates and empirically examine their effects on tabular foundation-model performance and rankings.
These examples show why a shared population optimum does not imply identical finite-sample learning behavior.
Our regression comparisons complement the classification experiments of \citet{merkle2013choosing}.
\item \textbf{Empirical validation: fine-tuning and cross-metric benchmarking.}  
These findings are validated by fine-tuning tabular foundation models with different scoring rules (CRLS \citep{juutilainen2012exceedance}, $\beta$-energy score, density power divergence scores, etc.) and reveal non-trivial interactions between training objectives and evaluation metrics across the regression datasets \citep{landsgesell2026scoringbench}. This indicates that training objectives can influence model behavior, and that practitioners can adapt foundation models to task-specific scoring objectives.
\end{itemize}

\section{Preliminaries}
\label{sec:preliminaries}

\subsection{Classical and distributional regression}

Given an input space $X \subseteq \R^d$ and an output space $Y \subseteq \R$,
classical (one-dimensional) regression aims at reconstructing a function $f: X \rightarrow Y$
from $n$ observations $\mathcal{D} := \{(\bm{x}_i, y_i)\}_{i=1}^n \in (X \times Y)^n$.
These observations $\mathcal{D}$ are usually assumed to be i.i.d.\ samples from an unknown probability distribution $P$ on $X \times Y$.
Classical regression models a single functional of the conditional distribution $P_{Y \mid \mathbf{X} = \mathbf{x}}$. 
A very common objective is the minimization of the mean squared error,
which is an example for a \textit{scoring function} \citep{gneiting2011making}:
\begin{equation}
    f^* = \text{argmin}_{f} \; \mathbb{E}\bigl[(Y - f(\mathbf{X}))^2\bigr],
    \label{eq:mse}
\end{equation}
whose minimizer is the \emph{conditional mean} $f^*(\mathbf{x}) = \mathbb{E}[Y \mid \mathbf{X} = \mathbf{x}]$ (see also Bayes estimator). 
Conversely, the conditional median arises under MAE minimization, and the $\tau$-quantile arises under the asymmetric pinball loss or the generalized piecewise linear form \citep{gneiting2011making, gneiting2023model}. In each case, the prediction problem reduces to estimating a single real number per input, namely a (conditional) \emph{point estimate}. 
While sufficient for many tasks, point estimates cannot capture \emph{aleatoric uncertainty}: 
the irreducible variability of $Y$ given $\bm{x}$ that persists even with infinite data \citep{KNEIB202399, wang2025aleatoric}.
A distinction must be drawn between this irreducible aleatoric uncertainty and epistemic uncertainty, which stems from limited data and can in principle be reduced.
A natural extension thereof is distributional regression,
which addresses this limitation by modeling the entire conditional distribution
$P_{Y | X=x}$ rather than a single functional thereof.
This can be done by estimating the conditional density, CDF, 
or multiple quantile functions \citep{KNEIB202399}.
This yields a full characterization of predictive uncertainty, 
from which point estimates, prediction intervals, and arbitrary quantiles can subsequently be derived.

\subsection{Calibration and Sharpness}
\label{sec:calibration_and_sharpness}

Calibration and sharpness are two complementary properties used to evaluate and compare probabilistic forecasts or predictive distributions,
as elaborated in \citet{gneiting2014probabilistic}.

\textbf{Calibration} refers to the statistical consistency between predictive distributions and observed outcomes: across many predictions, outcomes should behave as if drawn from the corresponding predictive distributions. 
Calibration encompasses several different concepts \citep{chidambaram2024reassessing}, including probability integral transform (PIT), T-, and threshold calibration \citep{gneiting2023regression, henzi2021isotonic}\footnote{In the classification setting, calibration is well investigated \cite{xia2026whatandwhat, manokhin2026classifier, berta2026calarena}. 
}.
Conformal predictive systems use "calibration" in the sense of marginal coverage guarantees \citep{tibshiraniforecast2023, allen2025sample}; this is, however, not directly linked to PIT calibration or other forms of calibration.

PIT calibration is commonly assessed with a histogram of probability integral transforms (PITs), as shown in \Cref{fig:pit}. For a predictive CDF $\hat{F}$, the PIT of an observed outcome $y$ is defined as follows \citep{gneiting2007calibrationandsharpness}:
\begin{equation}
    \label{eq:pit}
    u = \hat{F}(y)
\end{equation}
Under PIT calibration, these values are uniformly distributed on $[0,1]$. 
Typically, this is an aggregate property and does not establish whether individual forecasts are conditionally calibrated \citep{gneiting2023regression}.

\begin{figure}[htbp]
\centering
\pgfmathdeclarefunction{normalcdf}{2}{%
  \pgfmathparse{1/(1+exp(-0.07056*((#1-#2)/1)^3 - 1.5976*(#1-#2)/1))}%
}
\pgfmathdeclarefunction{normalpdf}{2}{%
  \pgfmathparse{exp(-((#1-#2)^2)/(2*1^2))/(1*sqrt(2*pi))}%
}

\begin{tikzpicture}[scale=0.85, transform shape]
 
\begin{axis}[
  name=leftpanel,
  width=3.2cm, height=7cm,
  xmin=0, xmax=1,
  ymin=0, ymax=1,
  axis lines=left,
  xtick=\empty,
  ytick={0,0.2,0.4,0.6,0.8,1.0},
  yticklabels={0,0.2,0.4,0.6,0.8,1.0},
  tick align=outside,
  ticklabel style={font=\small},
  enlargelimits=false,
  clip=false,
]
  \foreach \y in {0,0.0625,...,0.9376}{
    \addplot[fill=blue!20, draw=blue!30, line width=0.3pt]
      coordinates {(0,\y) (1,\y) (1,\y+0.0625) (0,\y+0.0625)} \closedcycle;
  }
\end{axis}
 
\begin{axis}[
  name=cdfpanel,
  at={(leftpanel.north east)}, anchor=north west,
  xshift=0.3cm,
  width=6cm, height=7cm,
  xmin=-3.5, xmax=3.5,
  ymin=0, ymax=1,
  axis lines=box,
  xtick=\empty,
  ytick={0,0.2,0.4,0.6,0.8,1.0},
  yticklabels=\empty,
  tick align=outside,
  enlargelimits=false,
  clip=true,
  legend style={draw=none},
]
  \addplot[blue!60!black, thick, smooth, domain=-3.5:3.5, samples=120]
    {normalcdf(x,0)};
 
  \node[anchor=north west, font=\small] at (axis cs:-3.4, 0.98)
    {$F_Y(y) = \mathbb{P}(Y \leq y)$};
 
  \foreach \yval/\pval in {-0.7/0.242, 0.3/0.618, 0.9/0.816, 1.5/0.933}{
    \addplot[gray, thin, dashed] coordinates {(\yval,0) (\yval,\pval)};
    \addplot[gray, thin, dashed] coordinates {(-3.5,\pval) (\yval,\pval)};
  }
\end{axis}
 
\begin{axis}[
  name=histpanel,
  at={(cdfpanel.south west)}, anchor=north west,
  yshift=-0.3cm,
  width=6cm, height=3.2cm,
  xmin=-3.5, xmax=3.5,
  ymin=0, ymax=0.45,
  axis lines=left,
  xtick=\empty,
  ytick=\empty,
  xlabel={$y$},
  xlabel style={font=\small, at={(axis description cs:1.02,0)}, anchor=west},
  enlargelimits=false,
  clip=true,
]
  \foreach \xi/\hi in {%
    -2.5/0.018, -2.0/0.054, -1.5/0.130, -1.0/0.242,
    -0.5/0.352, 0.0/0.399, 0.5/0.352,
     1.0/0.242,  1.5/0.130,  2.0/0.054,  2.5/0.018%
  }{
    \pgfmathsetmacro{\xlo}{\xi - 0.25}
    \pgfmathsetmacro{\xhi}{\xi + 0.25}
    \addplot[fill=blue!20, draw=blue!40, line width=0.4pt]
      coordinates {(\xlo,0)(\xlo,\hi)(\xhi,\hi)(\xhi,0)} \closedcycle;
  }
\end{axis}
 
 
\end{tikzpicture}
\caption{Illustration of the probability integral transform (PIT), used to assess the calibration of probabilistic predictions.}
\label{fig:pit}
\end{figure}

\textbf{Sharpness} refers to how concentrated or narrow predictive distributions are, e.g.\ measured by the width of the prediction intervals. 
The sharpness is independent of the observed outcomes. 
A forecast that always predicts a huge, vague interval can be perfectly calibrated (since the true value will almost always fall inside something so wide) but is nearly useless. 
Sharpness measures how much that uncertainty has been reduced,
i.e.\ narrower intervals or more peaked densities are sharper.
It can be measured for example by the width of prediction intervals \citep{gneiting2007calibrationandsharpness}, 
or by the standard deviation of the predictive distribution \citep{tran2020methods}.

Although the true conditional distribution $F(\,\cdot\mid\mathbf{x})$, the predictive distribution $\hat{F}$, and the observed outcome $y$ together form a triple $(F, \hat{F}, y)$, in practice only the pair $(\hat{F}, y)$ is available for evaluation \citep{gneiting2007calibrationandsharpness, dawid1984present}.
As \citet{gneiting2007calibrationandsharpness} notes, calibration is a property of this pair $(\hat{F}, y)$, while sharpness is a property of $\hat{F}$ alone.

Following \citet{gneiting2007calibrationandsharpness}, probabilistic forecasting seeks the sharpest possible forecasts subject to calibration: a calibrated forecast can still be uninformative if it lacks sharpness, while a sharp but uncalibrated forecast is not trustworthy.
In optimum score estimation, proper scoring rules serve as a unified optimization objective that incentivizes both calibration and sharpness.
A \emph{scoring rule} $S(\hat{F}, y)$ measures forecast quality by assigning a numerical score 
to the predictive distribution $\hat{F}$ given an observed outcome $y$, 
thereby unifying the evaluation of both calibration and sharpness in a single criterion \citep{gneiting2007strictly}.

\subsection{Proper scoring rules}

For a comprehensive introduction to scoring rules, we refer the reader to \citep{gneiting2007strictly, pic2025proper}.
In the following, we cover the basic definitions:
Let $\mathcal{Y} \subseteq \R$ be a set, $\mathcal{F}$ be a $\sigma$-algebra on $\mathcal{Y}$ and $\mathcal{P}$ be a set of probability measures on the measure space $(\mathcal{Y}, \mathcal{F})$.
Then a scoring rule is a function
\begin{align*}
S: \mathcal{P} \times \mathcal{Y} \rightarrow \overline{\R}, (\mathbb{P}, y) \mapsto S(\mathbb{P}, y),
\end{align*}
such that the integral $S(\mathbb{P}, \mathbb{Q}) := \mathbb{E}_{Y \sim \mathbb{Q}}[S(\mathbb{P}, Y)] =  \int S(\mathbb{P}, y) \, \mathrm{d}\mathbb{Q}(y)$ exists for all $\mathbb{P}, \mathbb{Q}$ and takes values in $\overline{\R}$.
A scoring rule $S$ is called \emph{proper} with respect to a class of probability distributions $\mathcal{P}$ if the expected score is minimized by the true distribution $\mathbb{Q}$, that is, 
\begin{align}
\label{eq:def_proper_scoring_rule}
\mathbb{E}_{y \sim \mathbb{Q}}[S(\mathbb{Q}, y)] \leq \mathbb{E}_{y \sim \mathbb{Q}}[S(\mathbb{P}, y)]
\end{align}
for all $\mathbb{P} \in \mathcal{P}$.
A scoring rule is \emph{strictly proper} if equality holds if and only if $P = Q$, 
ensuring that the true distribution is the unique minimizer and that no misspecified forecast can achieve the same expected score.

We furthermore recall some scoring rules, which will be used in the subsequent analysis and experiments.
First, the \textbf{CRPS} (continuous ranked probability score) which was introduced by \cite{matheson1976scoring} is given by
\begin{equation}
    \label{eq:crps}
    \text{CRPS}(\mathbb{P}, y) = \int_{-\infty}^{\infty} (F_{\mathbb{P}}(x) - \mathds{1}_{\{x \geq y\}})^2 \, \mathrm{d}x,
\end{equation}
where $F_{\mathbb{P}}$ is the cumulative distribution function (CDF) of $\mathbb{P}$,
see \Cref{fig:crps} for a visualization.
The CRPS weights all quantiles equally, which can be limiting when risk is asymmetric over the output space \citep{gneiting2011comparing}.
\cite{arnold2024decompositions} discusses CRPS decomposition into a miscalibration, discrimination and uncertainty component. 
Therefore, optimizing CRPS in principle targets all components (notably also the miscalibration component). \\
Second, we consider the \textbf{$\beta$-energy score} \citep{gneiting2007strictly}.
For any $\beta \in (0,2)$, it is defined as
\begin{align}
\label{eq:scor_beta_energy}
S_{\beta}(\mathbb{P}, y) = \mathbb{E}_{X \sim \mathbb{P}} \|X - y\|^\beta - \tfrac{1}{2}\, \mathbb{E}_{X, X' \sim \mathbb{P}} \|X - X'\|^\beta,
\end{align}
where $X, X' \sim \mathbb{P}$ are drawn independently from $\mathbb{P}$. 
The $\beta$-energy score is strictly proper on the space of probability distributions with finite $\beta$-moment, i.e.\ $\int \Vert x \Vert^\beta ~ \mathrm{d}\mathbb{P}(x) < \infty$.
For $\beta=1$ and a univariate probability distribution,
the $\beta$-energy score simplifies to the CRPS.
Note that for $\beta=2$, the expression simplifies to $\Vert \mathbb{E}[X] - y \Vert^2$,
i.e.\ the ordinary squared error of the mean forecast.
In this boundary case the score is no longer strictly proper, since the squared error elicits only the conditional mean rather than the full predictive distribution \citep{gneiting2011making}. \\
Third, the \textbf{CRLS} (Continuous Ranked Logarithmic Scoring Rule, also called the Exceedance Probability Score) \citep{juutilainen2012exceedance} is defined as
\begin{align}
\label{eq:scor_CRLS}
    \mathrm{CRLS}(\mathbb{P}, y) = -\int_{-\infty}^{\infty} \log |F_{\mathbb{P}}(x) + \mathds{1}\{y \le x\} - 1| ~ \mathrm{d}x,
\end{align}
where the additional negation ensures lower scores correspond to better forecasts.
CRLS combines the distance sensitivity of CRPS with a logarithmic penalty, giving more weight to events that were predicted to be very unlikely \citep{juutilainen2012exceedance}. 

Finally, the \textbf{Density Power Divergence (DPD)} family \citep{ghosh2013robust} is a parametric class of proper scoring rules indexed by $\alpha \geq 0$, defined as
\begin{align}
\label{eq:scor_dpd}
    S_\alpha(\mathbb{P}, y) = \int_{-\infty}^{\infty} f_{\mathbb{P}}(t)^{1+\alpha} \, \mathrm{d}t
    - \left(1 + \frac{1}{\alpha}\right) f_{\mathbb{P}}(y)^{\alpha}, \qquad \alpha > 0,
\end{align}
where $f_{\mathbb{P}}$ is the probability density of $\mathbb{P}$ and $y$ is the observed outcome.
Two familiar objectives are contained in this family.
Setting $\alpha=1$ in \eqref{eq:scor_dpd} yields the conditional density estimation (CDE) loss, the continuous-outcome counterpart of the Brier score \citep{rudemo1982empirical}.
The other connection arises as $\alpha \to 0$: after accounting for an additive constant, the objective becomes the negative logarithm of the predictive density, $S_0(\mathbb{P}, y) = -\log f_{\mathbb{P}}(y) + \mathrm{const}$.

\begin{figure}[htbp]
\centering
\input{Figures/vis_CRPS.tex}
\caption{Visualization of the CRPS from \Cref{eq:crps}. The shaded area indicates the pointwise discrepancy between the predictive CDF $F_{\mathbb{P}}$ and the Heaviside step function $\mathds{1}_{\{x \geq y\}}$ induced by the observation $y$; the CRPS is the integral of this squared difference over all thresholds $x$.}
\label{fig:crps}
\end{figure}

\section{Scoring rules for Probabilistic Forecasting: The Choice Matters}
\label{sec:choice}

The obvious drawbacks of using a mean value prediction despite having access to a full predictive output distribution have already been highlighted in \Cref{fig:x-shape}.
In particular, a ranking of distributional regression models based on mean values is therefore not a good choice
since the mean functional does not capture the full predictive distribution and therefore ignores aleatoric uncertainty.
Thus one should instead refer to scoring rules.
However, it is known that the particular choice of the scoring rules is not a trivial decision and may have measurable consequences for model performance and rankings \citep{merkle2013choosing,du2021beyond, waghmare2025proper}. 
Therefore, especially for small sample regimes,
the choice of the scoring rule for evaluation should be aligned with the given decision-making context \citep{carvalho2016overview}.
In asymmetric risk settings such as finance or weather forecasting, where over- and under-prediction carry unequal costs, this alignment can be realized through threshold- and quantile-weighted or otherwise asymmetric scoring rules \citep{gneiting2011comparing, iacopini2023proper}.

\begin{figure}[h]
    \centering
    \begin{subfigure}[t]{0.48\textwidth}
        \centering
        \includegraphics[width=\textwidth]{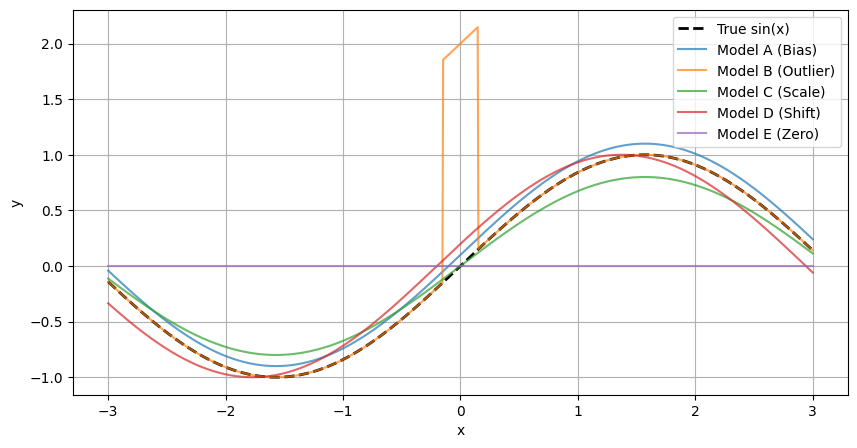}
        \caption{Toy models used for the analysis.}
        \label{fig:toy_models_decision}
    \end{subfigure}
    \hfill
    \begin{subfigure}[t]{0.48\textwidth}
        \centering
        \includegraphics[width=\textwidth]{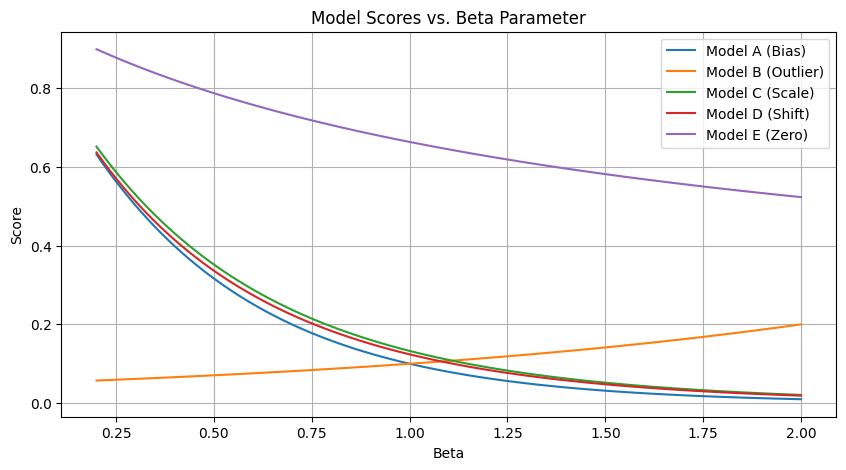}
        \caption{Model ranking as a function of the scoring rule parameter $\beta$.}
        \label{fig:toy_model_decision_scoring_rules}
    \end{subfigure}
    
    \caption{Effect of the choice of scoring rule on model evaluation. 
    Left: Synthetic toy models with distinct error characteristics. 
    Right: Corresponding model rankings under the $\beta$-energy score. 
    The preferred model depends on the choice of $\beta$: 
    Model~B performs best for $\beta \lesssim 1$, 
    while for $\beta \gtrsim 1$ the ranking changes, illustrating that different scoring rules induce different preferences over prediction errors.}
    
    \label{fig:toy_models_combined}
\end{figure}

\textbf{Example.}
In order to show how the choice of scoring rule affects the performance ranking of probabilistic models,
we consider a probabilistic forecast using highly concentrated predictive distributions of the following kind:
\begin{align}
\label{eq:concentrated_distributions}
\hat{p}_i(Y=y|X=x) = \delta_y(f_i(x)),
\end{align}
where $i\in\{A,B,C,D,E\}$ indexes different forecasting models (e.g. from different checkpoints of a training) 
and $\delta_z$ is the Dirac point measure at $z$.
The different models $i$ are described in \Cref{tab:models_and_ranks}, 
see \Cref{fig:toy_models_decision} for a plot.
Let us further impose the ground truth 
\begin{equation*}
p_g(Y=y|X=x) = \delta_y(g(x)).
\end{equation*}

\begin{table}[h]
\centering
\caption{Synthetic models and their rankings under different $\beta$-energy scoring rules for $x \in [-3,3]$ with $g(x)=\sin(x)$.}
\label{tab:models_and_ranks}
\begin{tabular}{l l l | c c c}
\toprule
\textbf{Model} & \textbf{Definition} & \textbf{Description} 
& \textbf{$\beta=0.2$} & \textbf{$\beta=1.073$} & \textbf{$\beta=2.0$} \\
\midrule
Model A & $f_A(x) = g(x) + 0.1$ 
& Small consistent bias 
& 2 & 1 & 1 \\

Model B & $f_B(x) = g(x) + 2\chi_{[-0.2, 0.2]}(x)$ 
& Large localized deviation 
& 1 & 2 & 4 \\

Model C & $f_C(x) = 0.8\, g(x)$ 
& Scaled version 
& 4 & 4 & 3 \\

Model D & $f_D(x) = g(x + 0.2)$ 
& Shifted version 
& 3 & 3 & 2 \\

Model E & $f_E(x) = 0$ 
& Trivial baseline 
& 5 & 5 & 5 \\

\bottomrule
\end{tabular}
\end{table}

\FloatBarrier

We evaluate the different models using the scale of $\beta$-energy scoring rules,
where a lower value indicates a better performance.
The resulting $\beta$-energy scores are depicted in \Cref{fig:toy_model_decision_scoring_rules},
and the resulting rankings of the models is listed in \Cref{tab:models_and_ranks}.
One can clearly observe that the ranking of the models depends on the value of $\beta$, 
i.e.\ on the considered $\beta$-energy scoring rule.
This behavior arises from the different weighting of the severity of errors depending on the scoring rule.
Some models make small errors everywhere (model A), 
some models make large errors in a small region (model B), 
some models make medium-sized errors everywhere (model C) and 
some models make small errors everywhere but are shifted (model D).

For Dirac forecasts as in \eqref{eq:concentrated_distributions},
the $\beta$-energy score formula \eqref{eq:scor_beta_energy} simplifies directly to familiar metrics: 
it reduces to the Mean Absolute Error (MAE) when $\beta=1$, 
and to the Mean Squared Error (MSE) when $\beta=2$. 
This connection illustrates how the choice of scoring rule ($\beta$) can affect model rankings.
Within the restricted class of Dirac forecasts, minimizing the expected $\beta$-energy score targets a conditional median when $\beta=1$ and the conditional mean when $\beta=2$ \citep{du2021beyond}.
For unrestricted distributional forecasts, CRPS instead targets the full conditional distribution, whereas the $\beta=2$ score identifies only its mean.

The example shows that selecting a score determines how errors of different sizes contribute to the comparison of forecasts \citep{waghmare2025proper}.
This dependence is compatible with propriety: the guarantee concerns an expectation under the data-generating distribution, whereas a fitted model is optimized and assessed using finitely many observations.
To evaluate distributional predictions throughout training, fine-tuning, and benchmarking of tabular foundation models, we therefore use the sample mean
\begin{align}
\hat{\mathbb{E}}_n\!\left[S(\hat{F}(\cdot \mid \mathbf{x}), y)\right] = \frac{1}{n} \sum_{i=1}^{n} S(\hat{F}(\cdot \mid \mathbf{x}_i), y_i),
\label{eq:empirical_score}
\end{align}
as the objective, replacing a loss such as MSE that evaluates only a point prediction.
The operator $\hat{\mathbb{E}}_n[\cdot]$ estimates the expectation by averaging over the $n$ observed pairs.
In gradient-based optimization, each update depends on $\nabla_{\hat{F}} \hat{\mathbb{E}}_n\!\left[S(\hat{F}(\cdot \mid \mathbf{x}), y)\right]$.
Changing the scoring rule $S$ can change the parameter gradients and hence the optimization trajectory.
With finite samples or a restricted model class, different strictly proper scoring rules can yield different fitted distributions and different rankings on held-out data.
The score therefore controls part of the learning procedure's inductive bias \citep{merkle2013choosing, waghmare2025proper, carvalho2016overview}.
The classification experiments of \citet{merkle2013choosing} illustrate this dependence empirically, while our regression example makes the differing evaluation preferences explicit through the $\beta$-energy family.

While this discussion demonstrates that scoring rule choice affects model training,
fine-tuning and benchmarking,
empirically determining which rule works best for a given task remains challenging 
and is explored in the following sections.

\section{Experiments and results}
\label{sec:results}

We use ScoringBench \citep{landsgesell2026scoringbench} as a testbed for studying how scoring-rule choice affects fine-tuning and evaluation, focusing on TabPFN.
We first examine alignment between fine-tuning objectives and evaluation performance, then complement this comparison with analyses of marginal coverage and correlations between scoring rules across the benchmark models.
For this work, we adapt the TabPFN fine-tuning code from GitHub, starting from the realTabPFNv2.5 checkpoint, and implement custom loss functions (the $\beta$-energy score, CRLS and others).
The datasets, preprocessing steps and hyperparameters for fine-tuning are described in \citep{landsgesell2026scoringbench}.

For the fine-tuning experiments,
we fine-tune base models using \eqref{eq:empirical_score} for various scoring rules
and subsequently evaluate these models using a comprehensive set of metrics spanning point-estimate accuracy ($\mathrm{MAE}$, $\mathrm{RMSE}$, $R^2$), 
proper scoring rules ($\mathrm{CRPS}$, $\beta$-energy scores, quantile weighted CRPS, $\mathrm{IS}_{\alpha}$, \dots), 
and additional diagnostics (coverage, sharpness, dispersion, training time). 
For descriptions of the metrics we refer to \cite{landsgesell2026scoringbench}.
Most metrics are lower-is-better; $R^2$ and dispersion are higher-is-better, marginal coverage should meet its nominal level.

\subsection{Fine-tuning objective alignment: Evaluation on fine-tuned vs.~other metrics}

\begin{figure}[!b]
\centering
\includegraphics[width=0.98\textwidth]{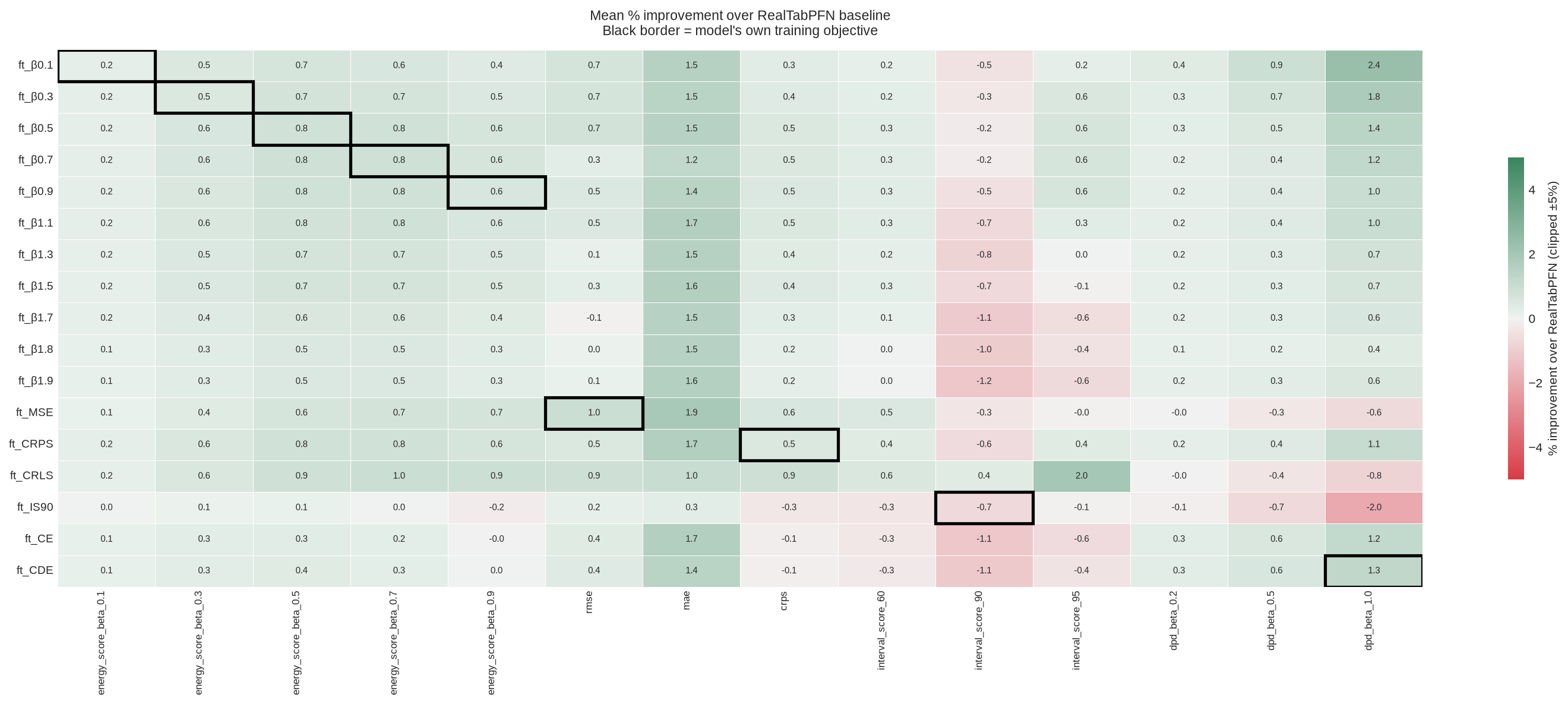}
\caption{Mean \% improvement over the RealTabPFN baseline (green = improvement, red = degradation, clipped at $\pm$30\%). Each row corresponds to a fine-tuned model; each column to an evaluation metric. Black borders mark a model's own training objective, enabling visual inspection of self-alignment (does fine-tuning on a given metric produce the best rank under that same metric?)}
\label{fig:fig2_self_alignment}
\end{figure}

\begin{figure}[!b]
\centering
\includegraphics[width=0.98\textwidth]{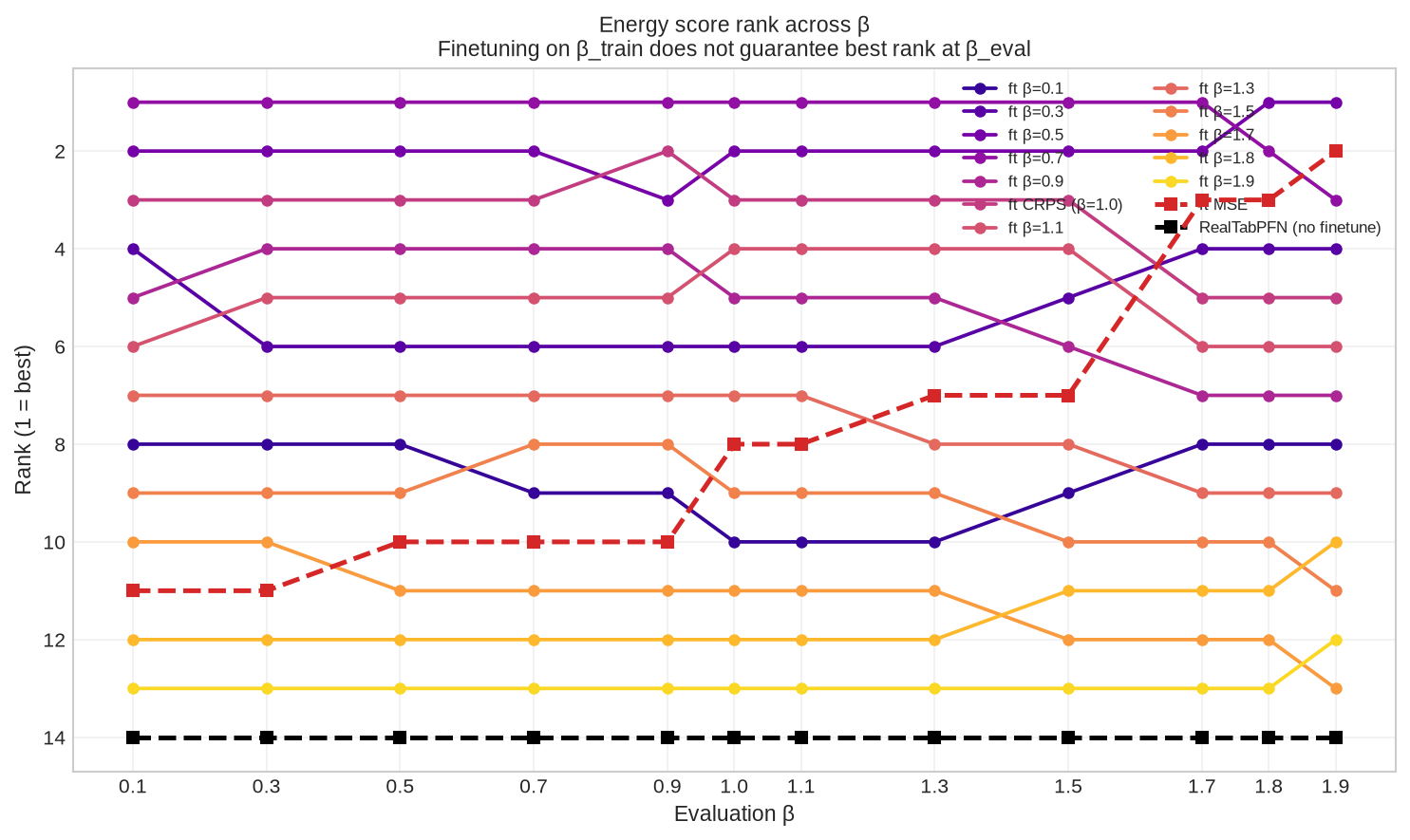}
\caption{Relative ranking of models under the $\beta$-energy score family.
For each evaluation $\beta$ on the $x$-axis, the models are ordered by their position in the full autorank ScoringBench leaderboard and then renumbered $1,\dots,N$ among the subset shown here, so the $y$-axis displays the \emph{relative} ranking within this subset (1 = best) rather than an absolute leaderboard rank.
Coloured solid lines are RealTabPFN fine-tuned on individual $\beta$-energy scores (CRPS corresponds to $\beta=1.0$); the dashed lines mark the MSE-fine-tuned variant and the non-fine-tuned RealTabPFN baseline (black).
Models fine-tuned at a given $\beta$ do not consistently achieve the best observed rank at that $\beta$.
In contrast, we observe that fine-tuning using $\beta=0.5$ ranks among the best models for all evaluated $\beta$ values.
}
\label{fig:fig4_energy_beta_profile}
\end{figure}

We first ask whether fine-tuning on a given objective translates into improved performance on the corresponding evaluation metric.
\Cref{fig:fig2_self_alignment} answers this across a broad set of objectives and metrics, showing the average improvement of several fine-tuned RealTabPFN variants over the unfinetuned baseline.
\Cref{fig:fig4_energy_beta_profile} then narrows the question to a single parametric family, the $\beta$-energy scores, where training and evaluation objectives differ only in the value of $\beta$.

In both cases, a model fine-tuned on a given objective does not necessarily rank best when evaluated on that same objective.
In \Cref{fig:fig4_energy_beta_profile}, for instance, the model fine-tuned with $\beta=0.5$ ranks among the best across the whole range of evaluation values of $\beta$, whereas models trained at larger $\beta$ do not consistently lead at their own value.
This contradicts the natural expectation that the training and evaluation objective should match, and it points to non-trivial interactions between the two.

A similar non-homogeneous effect emerges when the training objective is a point-estimate loss.
MSE fine-tuning reduces RMSE by approximately 1\% on average in this benchmark, while producing smaller gains or degradations under the displayed DPD scores (\Cref{fig:fig2_self_alignment}).

One possible explanation is that the objective interacts with the optimization budget, model capacity, and hyperparameters.
The present experiments do not isolate the contributions of these factors.
The scoring rule and its parameters thus act as a hyperparameter of the fine-tuning procedure.

\FloatBarrier

\subsection{Calibration}

As discussed in \Cref{sec:calibration_and_sharpness}, calibration encompasses multiple related notions, including marginal coverage and PIT uniformity \citep{gneiting2023regression}.
\Cref{sec:appendix_crepes} (\Cref{fig:coverage_amce}) addresses coverage calibration only, reporting the absolute marginal coverage error (AMCE) --- the coverage error $|\text{empirical} - \text{nominal}|$ computed per cross-validation fold and then averaged across folds and datasets --- at nominal levels from 20\% to 95\%; we use AMCE rather than plotting raw empirical coverage against nominal coverage because averaging signed coverage errors can mask systematic miscalibration through cancellation.

\subsection{Scoring rule correlation structure and scoring rule families}

Different proper scoring rules assess different aspects of forecast quality yet may produce correlated model scores in practice.
We compute Pearson correlations between scoring rules using per-dataset $z$-scores across all model--dataset pairs.
Correlations in scoring-rule performance broadly follow scoring-rule families, with density-based metrics correlating more strongly with each other than with energy-based scoring rules.
The heatmap and computational details are provided in \Cref{sec:appendix_corr_details} (\Cref{fig:metric_correlation}).

\FloatBarrier
\section{Conclusion}
\label{sec:conclusion}

Learning a predictive distribution requires attention to features that the accuracy of its mean cannot reveal.
For TabPFN and TabICL, much of the information in their outputs is therefore absent from assessments based predominantly on RMSE or $R^2$, as is common in tabular regression.
Proper scoring rules make this information part of the assessment by evaluating forecasts in terms of both calibration and sharpness.

Our main result concerns how these criteria shape the learning procedure itself.
Strict propriety identifies the true distribution as the population optimum, but a finite training sample does not fix the optimization trajectory independently of the objective.
The selected score can change the optimization updates and may therefore affect which predictive distributions are reached.
Our examples illustrate score-dependent rankings of fixed forecasts, while the foundation-model fine-tuning experiments document differences in predictive performance and leaderboard position across objectives.
The experiments also show that a model trained with a particular rule can be outperformed under that rule by a model trained with another objective.

These findings make objective selection a practical part of developing a tabular foundation model, alongside choices of architecture, optimizer, and regularization.
Deciding whether to fine-tune is only one step; practitioners must also consider which probabilistic objective reflects the intended decision problem.
Choosing that objective offers a way to adapt a pretrained model to the requirements of a particular task, while the evaluation rule specifies how its forecasts will be judged.

Benchmark reports should consequently assess distributional forecasts with proper scoring rules; our concurrent ScoringBench work provides a framework for doing so \citep{landsgesell2026scoringbench}.
Using such assessments together with probabilistic training objectives places the predictive distribution at the center of model development and connects its evaluation to the rest of the machine learning pipeline.

\subsection*{Limitations}

Small differences in expected scores need not imply similar asymptotic tail properties.
Certain tail functionals are not elicitable under the conditions studied by \citet{brehmer2019scoring}, while evaluating rare events can require large samples even when suitable scores are available \citep{lerch2017forecaster}.

The analysis and experiments in this paper concern a single response variable.
The approach extends to multiple targets by choosing multivariate scoring rules, including the variogram score.
\citet{pic2025proper} reviews suitable approaches.

\section*{Acknowledgement}
We thank David Holzmüller for a very engaging and supportive email exchange in March 2026.
We thank Noah Hollmann for a very polite and engaging call just a few days prior to Christmas 2025.
J.L. thanks Tobias Schröder, Lennart Purucker, and Rafael Izbicki for helpful discussions on density-based scoring rules.
T.W. gratefully acknowledges funding by Daimler and Benz Foundation as part of the scholarship program for junior professors and postdoctoral researchers.

\section*{Impact Statement}
This paper presents work whose goal is to advance the field of distributional regression in tabular foundation models. 
Besides use-case specific optimizations, there are no societal consequences of our work which we feel must be specifically highlighted here.

\bibliographystyle{unsrtnat}
\bibliography{references} 

\appendix

\section{Marginal Coverage and Conformal Predictive Systems with CREPES}
\label{sec:appendix_crepes}
\label{sec:appendix_crepes_variants}

Conformal predictive systems guarantee marginal coverage: prediction intervals derived from their predictive distributions have the correct nominal coverage in expectation \citep{allen2025sample}.
However, this does not imply better PIT calibration \citep{allen2025sample, gneiting2007calibrationandsharpness}.
The \texttt{crepes} package \citep{bostrom2022crepes} provides conformal prediction methods and we consider these three variants:
\begin{itemize}
  \item \textbf{Standard}: prediction intervals are independent of the covariates, i.e.\ dispersion \citep{tran2020methods} is absent.
  \item \textbf{Difficulty-controlled}: intervals adapt to the covariates via a ``difficulty estimator''; however, non-informative difficulty estimators can perform worse than the standard variant \citep{bostrom2022crepes}.
  \item \textbf{Mondrian}: the data are blocked into covariate-defined categories and the standard system is applied within each block, which is effective for heteroscedastic residuals \citep{bostrom2022crepes}.
\end{itemize}
Due to the covariate-independence of the standard variant, we also add the difficulty-controlled and Mondrian variants in the experiments.
Not much connection has been made between the conformal prediction literature and the broader forecasting literature on proper scoring rules \citep{tibshiraniforecast2023, allen2025sample}.

\begin{figure}[htbp]
\centering
\includegraphics[width=0.7\textwidth]{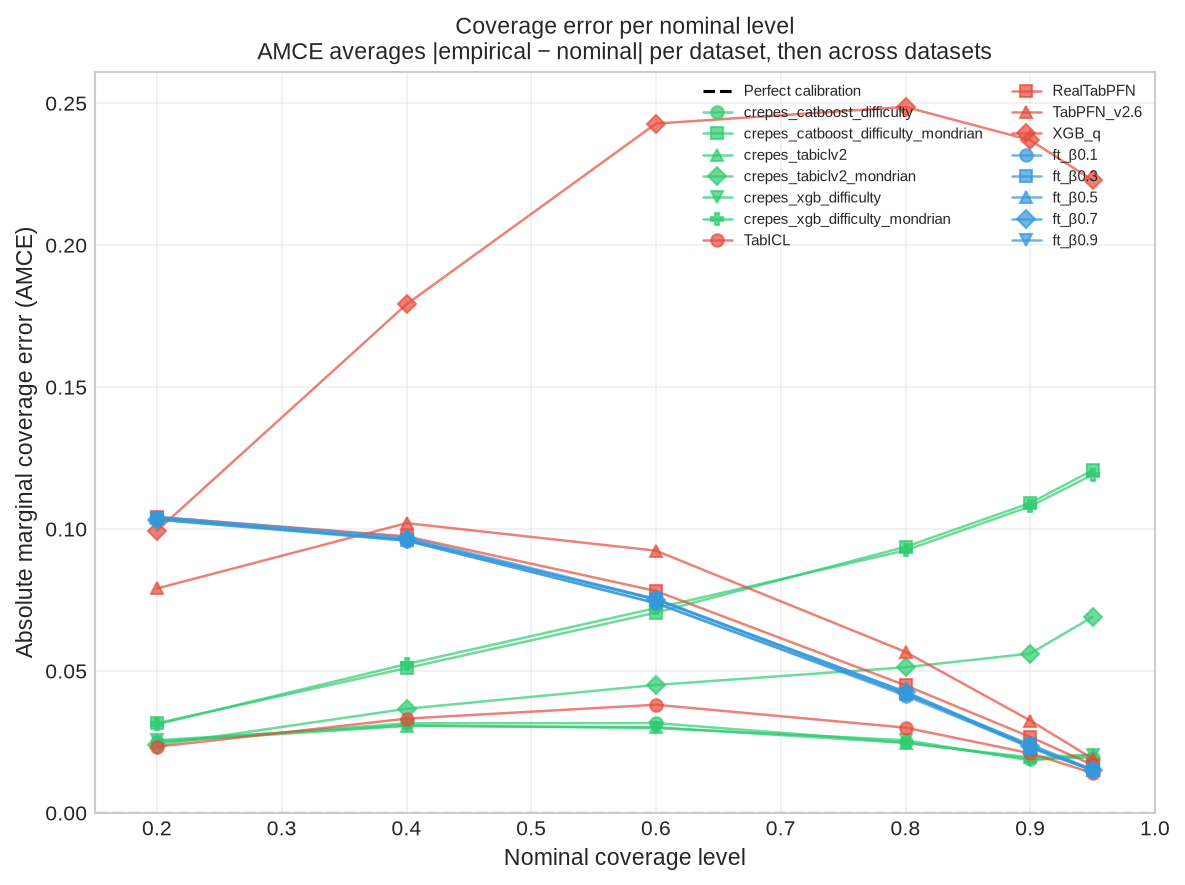}
\caption{Absolute marginal coverage error (AMCE) at nominal levels 20\%, 40\%, 60\%, 80\%, 90\%, and 95\%. For each model, dataset, and cross-validation fold, $|\text{empirical coverage} - \text{nominal level}|$ is computed first; these per-fold absolute errors are then averaged across folds within a dataset and across datasets, so over- and under-coverage cannot cancel out at either the fold or the dataset level, unlike when averaging empirical coverage directly; perfect calibration corresponds to the zero line. Each line represents one model; colour groups distinguish CREPES (green), Baseline/Other (red), and Fine-tuned (blue) models, with distinct marker shapes for identifiability within each group. The CREPES curves correspond to the difficulty-controlled and Mondrian variants of the conformal predictive systems from \texttt{crepes} \citep{bostrom2022crepes}. The difficulty-controlled CREPES variants maintain low coverage error across all nominal levels, whereas the Mondrian variants degrade sharply at higher nominal levels; among the non-conformal models, TabICL also stays close to perfect calibration throughout, while the remaining baseline models show large coverage errors, pronounced around the 60\% nominal level.}
\label{fig:coverage_amce}
\end{figure}

\Cref{fig:coverage_amce} shows the absolute marginal coverage error (AMCE) at each nominal level for the benchmark models: for each model, dataset, and cross-validation fold we compute $|\text{empirical coverage} - \text{nominal level}|$, then average these per-fold absolute errors across folds within a dataset, and finally across datasets. We report AMCE rather than plotting raw empirical coverage against nominal coverage because averaging signed coverage errors can hide systematic over- and under-coverage through cancellation; taking the absolute value on each fold before any averaging rules out both fold-level and cross-dataset cancellation while also measuring variability of the results.

\FloatBarrier

\begin{figure}[htbp]
\centering
\begin{subfigure}[t]{0.48\textwidth}
  \centering
  \includegraphics[width=\textwidth]{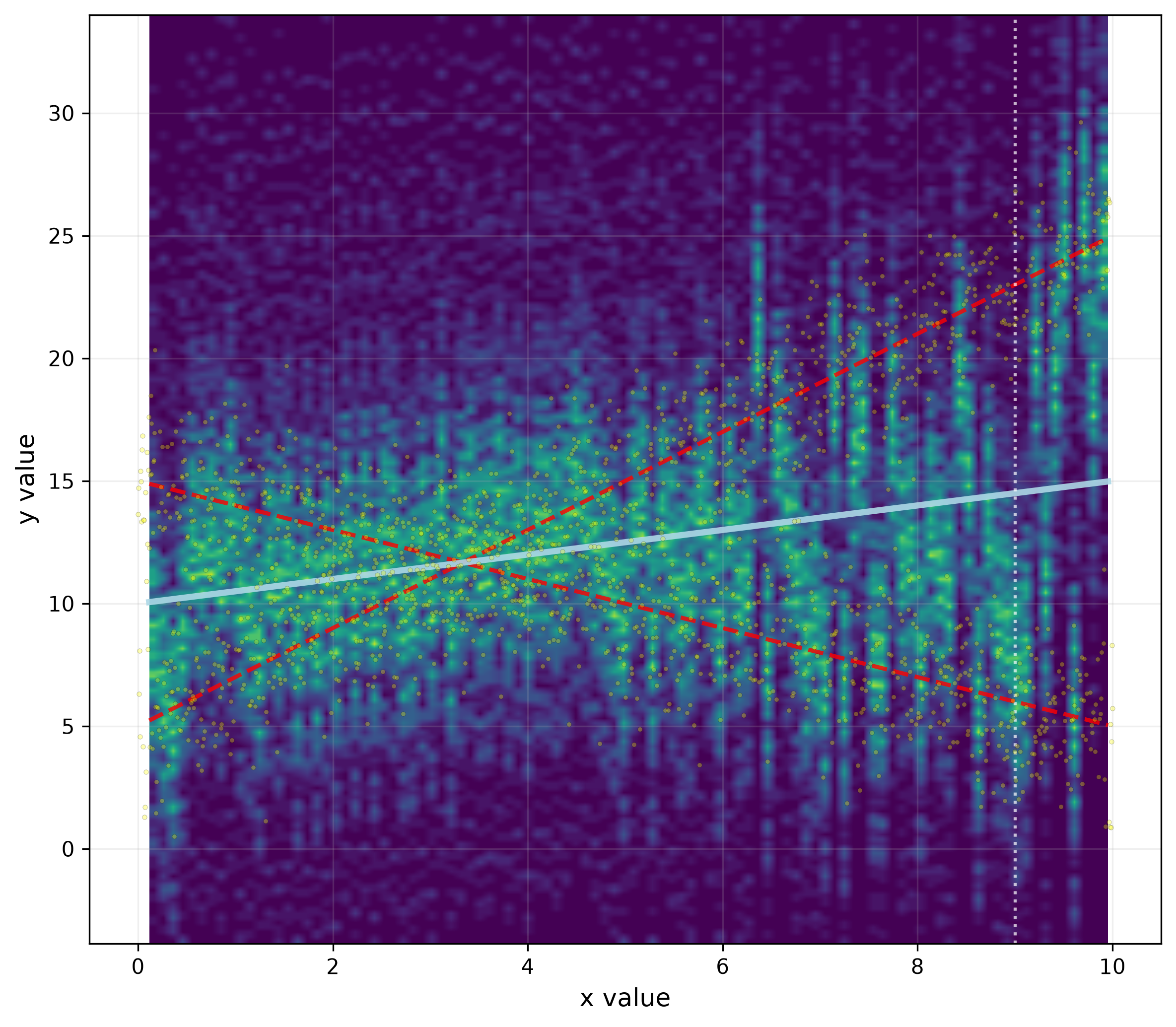}
  \caption{CREPES Standard (no difficulty estimator, no Mondrian).}
  \label{fig:x-shape-crepes-standard}
\end{subfigure}
\hfill
\begin{subfigure}[t]{0.48\textwidth}
  \centering
  \includegraphics[width=\textwidth]{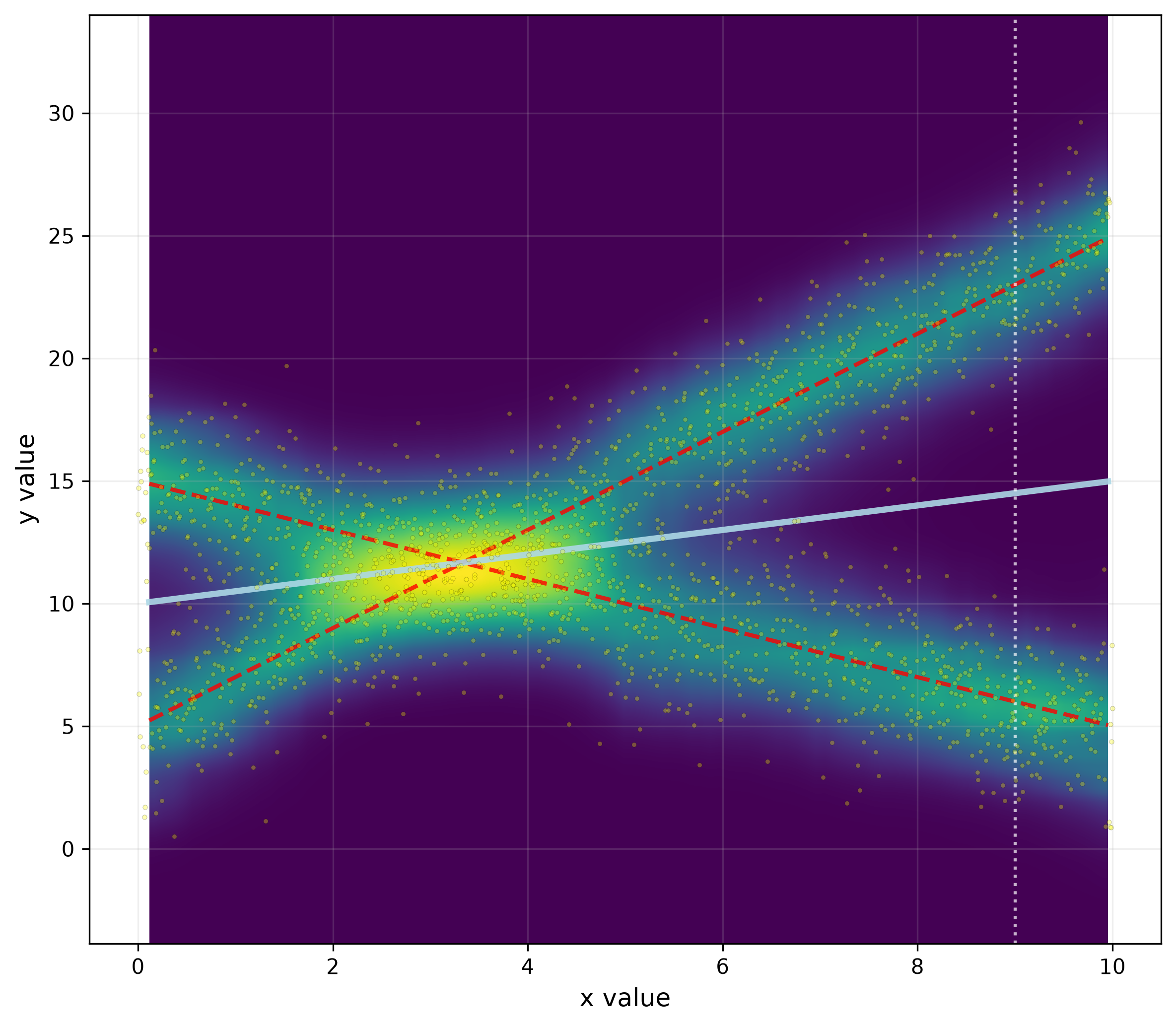}
  \caption{TabPFNv2.5 Baseline (shown in place of the unsupported Mondrian-only variant).}
  \label{fig:x-shape-crepes-tabpfn}
\end{subfigure}

\vspace{0.5em}

\begin{subfigure}[t]{0.48\textwidth}
  \centering
  \includegraphics[width=\textwidth]{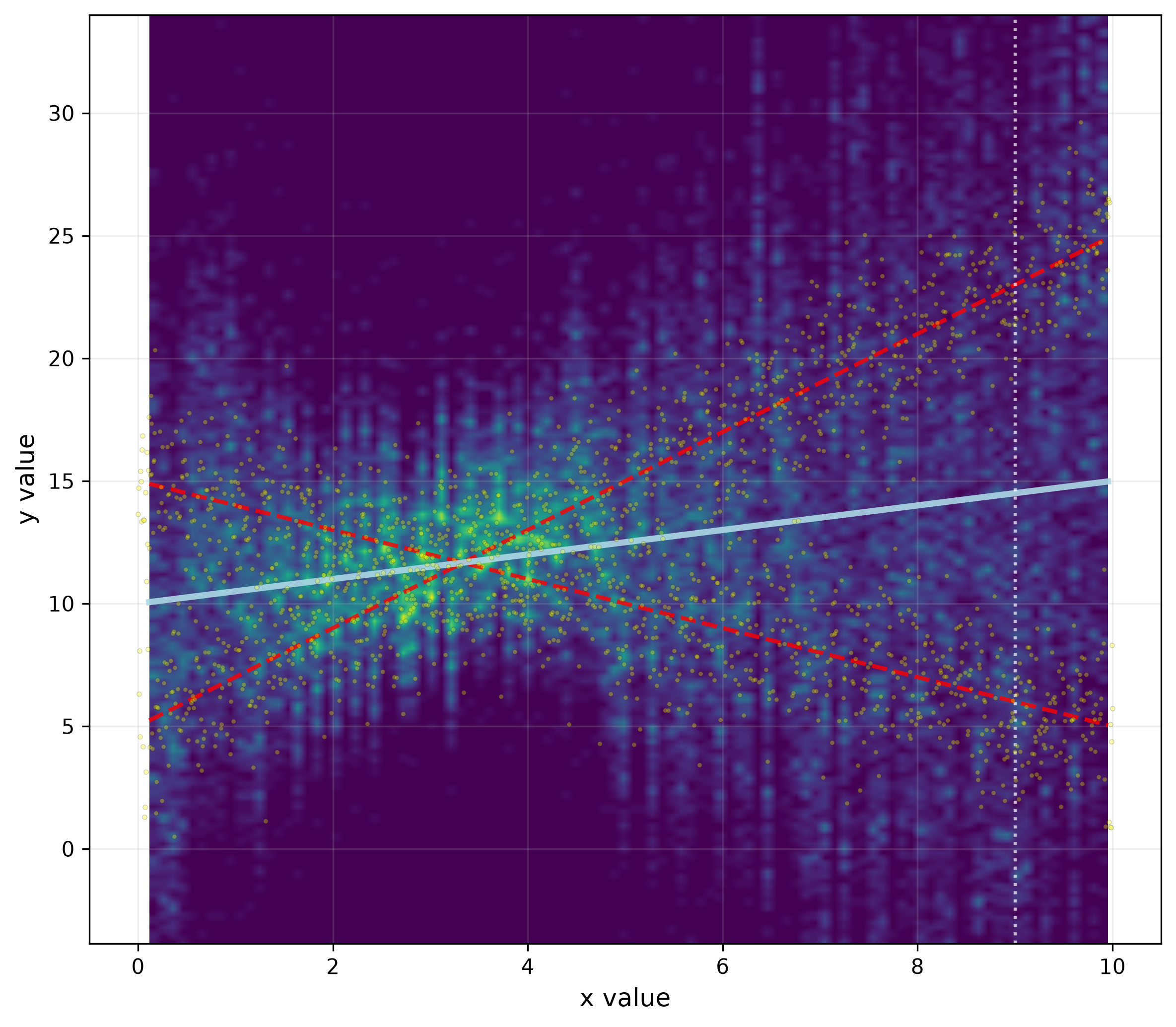}
  \caption{CREPES Difficulty-controlled (with difficulty estimator).}
  \label{fig:x-shape-crepes-normalized}
\end{subfigure}
\hfill
\begin{subfigure}[t]{0.48\textwidth}
  \centering
  \includegraphics[width=\textwidth]{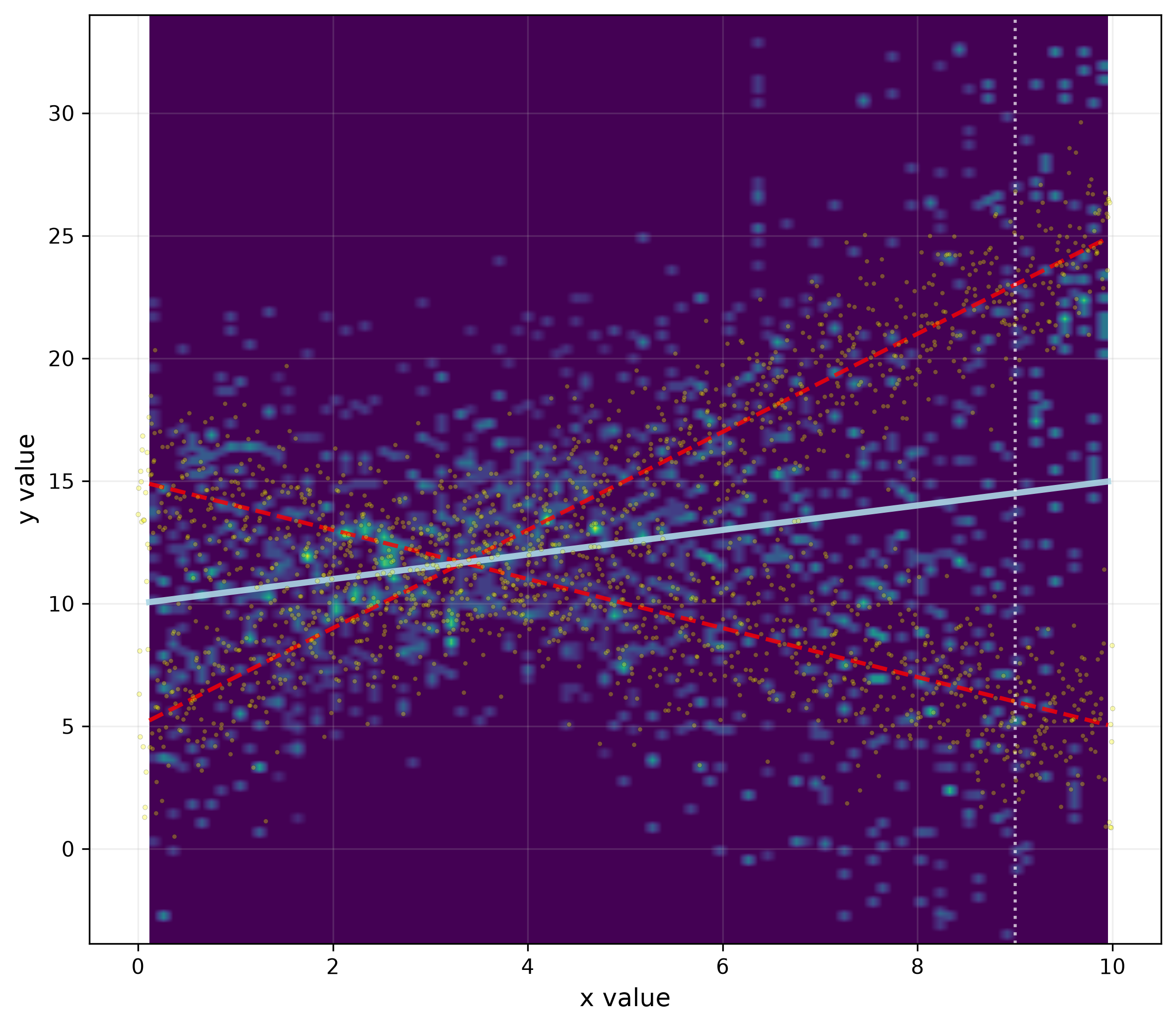}
  \caption{CREPES Difficulty-controlled Mondrian (difficulty estimator and Mondrian categorization).}
  \label{fig:x-shape-crepes-mondrian}
\end{subfigure}

\caption{Predicted probability densities (color) for the bimodal toy dataset from \Cref{fig:x-shape},
evaluated across four configurations of CREPES and the TabPFNv2.5 baseline.
Red dashed lines indicate the mean values of the two superimposed ground-truth generating processes;
the light blue line shows the conditional mean.
All three CREPES variants fail to capture the bimodal structure of the distribution,
whereas TabPFNv2.5 (\Cref{fig:x-shape}) recovers it successfully.}
\label{fig:x-shape-crepes}
\end{figure}

In analogy to \Cref{fig:x-shape}, which illustrates the distributional predictions of TabPFNv2.5 on the bimodal toy dataset, \Cref{fig:x-shape-crepes} below applies the same evaluation to the configurations of CREPES: standard (no difficulty estimator, no Mondrian), difficulty-controlled (with difficulty estimator), Mondrian, and difficulty-controlled Mondrian.
While TabPFNv2.5 captures the bimodal structure of the data-generating process, all CREPES variants fail to recover the bimodal distribution of the data generating process.
This illustrates a fundamental limitation of the used split conformal predictive systems which require that the residuals $R$ are independent of $X$.
However in the bimodal toy dataset in \Cref{fig:x-shape-crepes} the residuals are not independent of $X$,
i.e.\ $R \not \perp X$.

\FloatBarrier

\section{Details: Scoring Rule Correlation Analysis}
\label{sec:appendix_corr_details}

We first aggregate fold-level scores to one value $\bar{s}_{m,d,p}$ per combination of model~$m$, dataset~$d$, and scoring rule (metric)~$p$.
We then compute the per-dataset $z$-score
\begin{equation}
\label{eq:zscore_corr}
    z_{m,d,p} = \frac{\bar{s}_{m,d,p} - \mu_{d,p}}{\sigma_{d,p} + \epsilon}\,,
\end{equation}
where $\mu_{d,p} = \frac{1}{M}\sum_{m=1}^{M}\bar{s}_{m,d,p}$ is the mean of metric~$p$ across all $M$ models on dataset~$d$, $\sigma_{d,p} = \sqrt{\frac{1}{M-1}\sum_{m=1}^{M}(\bar{s}_{m,d,p}-\mu_{d,p})^2}$ the corresponding cross-model sample standard deviation, and $\epsilon = 10^{-9}$ prevents division by zero (following \cite[Appendix~D.2]{landsgesell2026scoringbench}).
The resulting $z_{m,d,p}$ measures how far model~$m$ lies from the average performance on dataset~$d$ and metric~$p$, expressed in units of the spread across models.
The correlation analysis thus compares relative deviations within each dataset rather than absolute score levels, which can differ markedly with dataset difficulty and metric scale.
We then compute the Pearson correlations between pairs of scoring rules $p$ and $p'$ across all model--dataset pairs $(m,d)$:
\begin{equation}
\label{eq:pearson_corr}
    \mathrm{Corr}(z_p,\, z_{p'})
    = \frac{\displaystyle\sum_{m=1}^{M}\sum_{d=1}^{D}\bigl(z_{m,d,p}-\bar{z}_p\bigr)\bigl(z_{m,d,p'}-\bar{z}_{p'}\bigr)}
           {\sqrt{\displaystyle\sum_{m=1}^{M}\sum_{d=1}^{D}\bigl(z_{m,d,p}-\bar{z}_p\bigr)^2\;\cdot\;\sum_{m=1}^{M}\sum_{d=1}^{D}\bigl(z_{m,d,p'}-\bar{z}_{p'}\bigr)^2}}\,,
\end{equation}
where $\bar{z}_p = \frac{1}{MD}\sum_{m=1}^{M}\sum_{d=1}^{D} z_{m,d,p}$ is the grand mean of the $z$-scores for metric~$p$ over all $M$ models and $D$ datasets.

\begin{figure}[htbp]
\centering
\includegraphics[width=0.95\textwidth]{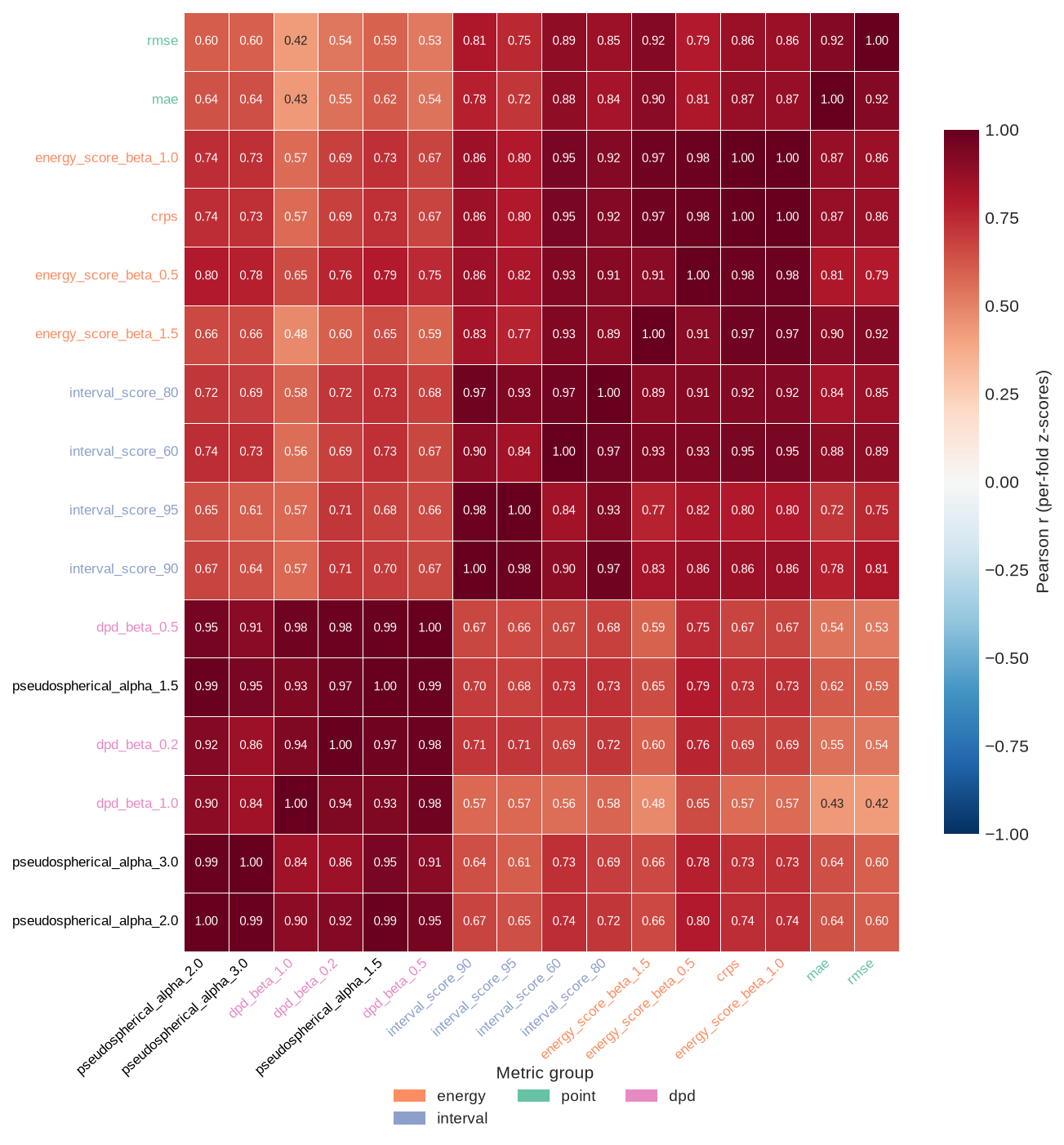}
\caption{Heatmap visualization of scoring rule correlations
via the Pearson correlation matrix (computed across all model--dataset pairs).
Label colours indicate metric families (see legend).
We observe that the density based metrics correlate more strongly with each other than with the energy based scoring rules.}
\label{fig:metric_correlation}
\end{figure}

\Cref{fig:metric_correlation} visualizes the resulting correlation matrix.

\end{document}